\documentclass[journal]{IEEEtran}
\usepackage[utf8]{inputenc}
\usepackage{amsmath, amssymb}
\usepackage{colortbl}
\usepackage{scalerel}
\usepackage[ruled,vlined]{algorithm2e}
\usepackage{multicol,tabularx,capt-of}

\begin{document}
\title{\LARGE{\bf ENHANCED SINGLE-SHOT DETECTOR FOR SMALL OBJECT DETECTION IN REMOTE SENSING IMAGES}}
\author{Pourya Shamsolmoali, Masoumeh Zareapoor, Eric Granger, Jocelyn Chanussot, Jie Yang

\thanks{\hspace*{-1.8em}P. Shamsolmoali, M.Zareapoor, J.Yang are with School of Automation, Shanghai Jiao Tong University, Shanghai, China, \{pshams, mzarea, jieyang\}@sjtu.edu.cn}
\thanks{\hspace*{-1.8em}E. Granger is with Dept. of Systems Engineering, \'Ecole de technologie sup\'erieure, Universit\'e du Qu\'ebec, Canada, Eric.Granger@etsmtl.ca}
\thanks{\hspace*{-1.8em}J.Chanussot is with Univ. Grenoble Alpes, INRIA, CNRS, Grenoble INP, LJK, Grenoble, France.}
\thanks{This article has supplementary material provided by the authors and color versions of one or more figures available at
https://doi.org/XXXXX.}
}


\markboth{42 IEEE International Geoscience and Remote Sensing Symposium (IGARSS)}%
{Shell \MakeLowercase{\textit{et al.}}: Bare Advanced Demo of IEEEtran.cls for IEEE Computer Society Journals}

\maketitle

\begin{abstract}
Small-object detection is a challenging problem. In the last few years, the convolution neural networks methods have been achieved considerable progress. However, the current detectors struggle with effective features extraction for small-scale objects. To address this challenge, we propose image pyramid single-shot detector (IPSSD). In IPSSD, single-shot detector is adopted combined with an image pyramid network to extract semantically strong  features for generating candidate regions. The proposed network can enhance the small-scale features from a feature pyramid network. We evaluated the performance of the proposed model on two public datasets and the results show the superior performance of our model compared to the other state-of-the-art object detectors.
\end{abstract}
\section{Introduction}
\label{sec:intro}

With the rapid progress in remote sensing technology, the analysis of remote sensing imagery (RSI) is a popular field because of its impact in both academic and industry. Object detection in RSI is an essential research field and has been studied and to address the practical issues, several object detection methods \cite{han2021multi, shamsolmoali2021multipatch, xu2021assd} have been developed.
In the last few years, object detection in natural images has been highly successful due to the great progress of deep learning models, including single-shot multibox detectors (SSDs) \cite{liu2016ssd}, models for region-based convolutional neural networks (R-CNNs) \cite{girshick2014rich, girshick2015fast, ren2015faster}, feature pyramid network for object detection (FPN) \cite{lin2017feature}, and you only look once (YOLO) models \cite{redmon2016you, redmon2018yolov3}. Recently the object detection methods that are used for natural images have been applied for object detection in RSI \cite{dong2019sig, ming2021cfc, wang2019fmssd, yuan2021olcn, shamsolmoali2021rotation}. Dong et al. \cite{dong2019sig} proposed a model based on \cite{ren2015faster} and adopted transfer learning to reduce the possibility of missing small objects. In \cite{ming2021cfc}, the authors proposed a feature capturing network based on FPN to enhance detection accuracy by improving feature representation, and optimizing label assignment. Wang et al. \cite{wang2019fmssd} introduced an architecture named feature-merged single-shot detection network (FMSSD), which integrates the information of various sizes by using FPN and different atrous rates to enhance the quality of features. In \cite{yuan2021olcn}, a model is proposed for small object detection by using low coupling regression and receptive field optimizing layer for better estimation of Regions of Interests (RoIs).
In \cite{shamsolmoali2021rotation}, the authors proposed an architecture to extract semantically strong features in various scales and orientations for better small object detection in RSI. 
Nevertheless, the problem of small object detection is neglected in the existing models and there is a considerable scope to improve the models' performance. As we earlier discussed, it is challenging to accurately detect small objects that only occupy the region of $10 \times 10$ pixels in RSI. In this paper, we propose a new architecture on the basis of SSD to address the above challenges. The main contributions of this paper are as follows.\\

\begin{itemize}
\item{ We devise a detection pipeline for small objects by integrating an image pyramid network into SSD (IPSSD) to achieve more strong semantic features.}
\item{ We propose the rotation pooling layer to cover both the horizontal and oriented region proposals and design a tailored feature fusion model to make the extracted features fuse in a better form.}
\item{ We evaluate several recent object detection models in RSIs and the performances are stated.}
\end{itemize}

\begin{figure*}
  \centering
  \includegraphics[width=1\textwidth]{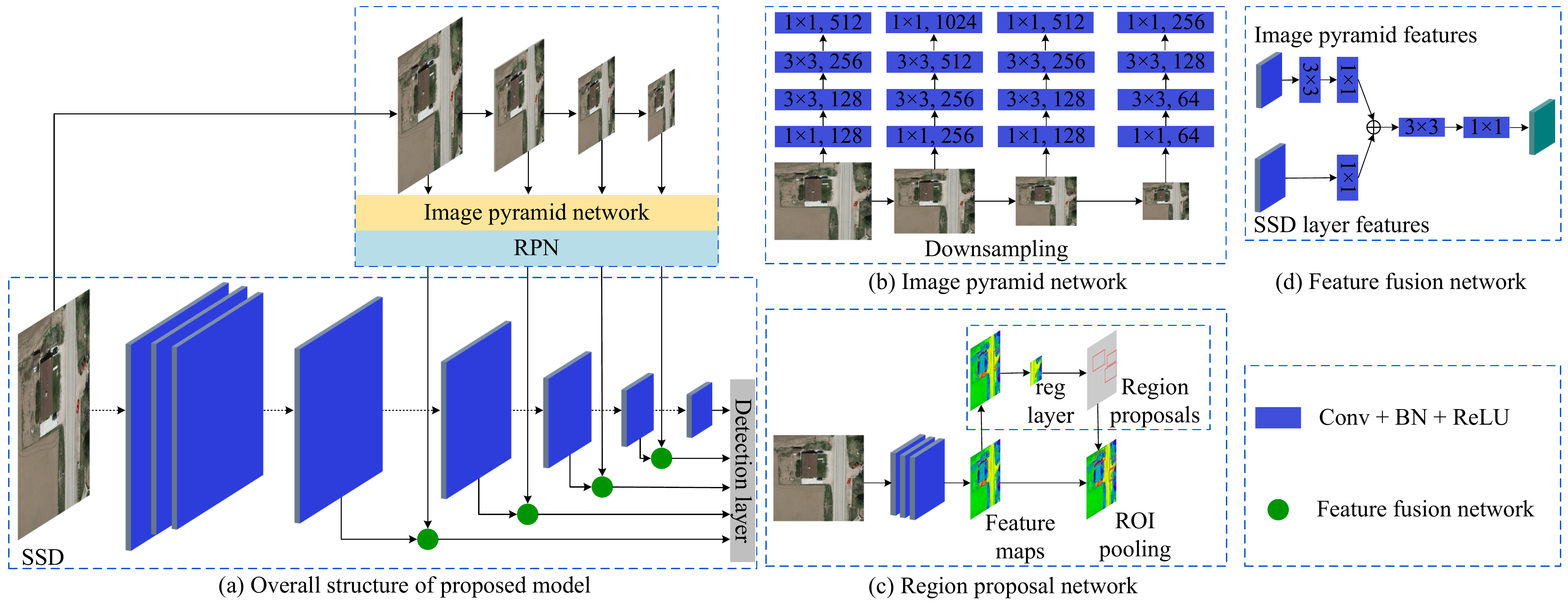}
\caption{Pipeline of IPSSD. (a) Network architecture. SSD is adopted with (b) the image pyramid network to extract candidate regions. (c) and (d) shows the RPN and the FFN architecture respectively.}
\label{fig:1}
\end{figure*}
\section{METHODOLOGY}
\label{sec:format}

The SSD \cite{liu2016ssd}, has shown a promising detection results. In SSD, each prediction layer has different resolutions, where the shallower layers participate in small targets detection and the deeper layers are contributing in large targets prediction. Despite of its high performance, SSD can not detect small objects due to the poor semantic information in earlier layers of the SSD. To address this problem, we enhance the features maps quality by integrating SSD with our propose image pyramid network (IPN) to extract ROIs. Different from the max-pooling layer in the region proposal network (RPN) that only able to cover the horizontal region proposals, the propose rotation pooling layer can handle both the horizontal and oriented region proposals. Moreover, a feature fusion network (FFN) is devised to improve the context information. Fig. \ref{fig:1} illustrates the architecture of IPSSD.\\

In our propose architecture, the SSD is used as the baseline detector, in which each layer detects a specific scale objects. This implies, the shallower layers estimate small objects, while large objects are estimated by deeper layers. However, due to insufficient semantic information in the shallower layers the SSD can not accurately detect small targets. To solve this problem, we extend the SSD with the IPN to enhance the SSD's  performance.
As showed in Fig. \ref{fig:1}, IPSSD contains two main streams: the standard SSD plus the IPN. For SSD, the backbone is VGG-16 and smaller convolution layers are added for better feature extraction. In our model, layers of IPN in different scales are integrated into the SSD's layers using FFN.
\subsection{Image pyramid network}
\label{ssec:subhead}
The standard FPNs \cite{lin2017feature} are not computationally efficient and efficient as various scales of each image is processed by a CNN. To handle this problem, we propose an effective model to generate object candidates through RPN in IPN \cite{ren2015faster}. The network contains a down-scaling process. The IPN as input receives different size images to generate a set of box offsets. Then according to scales of box offset, the module picks a feature map in the optimum size. Firstly form input image $X$, the model generates multi-scale image $X_p=\{x_1, x_2, ..., x_n\}$ by down-scaling the input image $X$, where, $n$ denotes the layers of IPN. To build multi-scale feature maps, the images are processed by the IPN $S_p=\{s_1, s_2, ..., s_n\}$, where, $S_p$ denotes the features of each layer. The IPN has two $1\times1$ and two $3\times 3$ $Conv$ layers with different number of channels.

\subsection{Oriented candidate regions network}
\label{sec:subhead}

The standard RPN takes the anchor to create the ROIs. Nevertheless, in RSI, the objects have a tiny scale with various orientations. Indeed, the horizontal candidates created by the standard RPN are not sufficient for challenging objects in the RSIs. To address this problem, we modify the standard RPN as follows: 1) We deleted the last three fully connected and soft-max layers; 2) a network called $reg-conv$ is added before the convolutional layer $conv[5-3]$; 3) a convolution kernels with a size of $3 \times 3 \times 512$ is employed to generate the 512D feature vector on the categorised feature maps; 4) the generated feature vector is processed by the $pred-score$ and $pred-bbox$ layers. For the oriented anchor scheme we followed \cite{dong2019sig} to create ROIs with various orientations and generate more suitable regions for a better small target detection. More specific, the candidate region $H\times W$ is splitted into several sub-regions. Thus, the sub-regions have the equivalent orientation as of the candidate region and each sub-region has size of $S_h\leftarrow \frac{h}{H}$, $S_w\leftarrow \frac{w}{W}$. In our model the rotation region proposal for each input is defined by $(x,y,h,w,\theta)$, in which $(x, y)$ is the centre of bounding box, $(h, w)$ are the height and width of bounding box (bbox) respectively, and $\theta$ denotes the standpoint from the absolute direction of the $x$-axis to the long side of the oriented bbox with a spatial size $S$. Therefore, the upper left corner of each sub-region is calculated as:
\begin{equation}
x_0,\;  y_0\leftarrow x - \frac{h}{2} + u S_h, \; \; y - \frac{w}{2} + v S_w
\end{equation}
in which $u\in\{0,1,...,H-1\}, v\in\{0,1,...,W-1\}$ and the rotated coordinate of $(x_0,y_0)$ computed as follows:

\begin{equation}
\begin{aligned}
\acute x\leftarrow(x_0 - x)cos\theta + (y_0 - y)sin\theta + x \\
\acute y\leftarrow(y_0 - y)cos\theta + (x_0 - x)sin\theta + y
\end{aligned}
\end{equation}

\subsection{Feature fusion network}
\label{sec:subhead}

To improve the spatial information, we introduce the FFN to combine features from the IPN layers with the SSD layers (see Fig. 2(d)). In the FFN, first, the output of IPN layer goes through a $3 \times 3$ and $1 \times 1$ conv layers, however, the output of each SSD layer only goes through a $1 \times 1$ conv layer. Then, the features of each IPN layer $l_{n-1}$ and SSD layer $l_n$ are combined via addition. Then, there are a $3\times3$ and $1\times1$ $Conv$ layers for detection $d_n= R (\eta(l_{n-1})\oplus \eta_n (l_n))$, where, $\eta_n(.)$ denotes the processes including a $1\times1, 3\times3$ $Conv$, BN layers, and $R$ represents the ReLU activation.

\section{EXPERIMENTS}
\label{sec:pagestyle}

In this section, the performance of our model is evaluated compare to the other approaches on 3 classes of the DOTA \cite{xia2018dota} and NWPU VHR-10 \cite{dong2019sig} datasets for small object detection. 
DOTA and NWPU VHR-10  are two large RSI dataset use for object detection and contains 15 and 10 classes of objects respectively with various scales and orientations.\\
For training and testing phase, the images are divided into the $600 \times 600$ pixels patches with an overlap of 100 pixels. Since, the number of images are not enough for training, to increase the number of images in training set, we apply rotation, and rescaling. All our experiments were conducted on a Tesla P-40 GPU. Our model is implemented using Pytorch.
The model uses VGG-16 as the backbone which pre-trained on the ImageNet and fine-tuned on RSI dataset. The SSD uses $(Conv4)$ and the fully connected layer ($FC$) layers of VGG-16 is transformed to a $Conv$ layers. In SSD the last FC layer of the VGG-16  is changed to several small ranges $Conv$ layers: $[Conv8,..., Conv11]$, with various feature sizes. The layers of SSD are merged with their corresponding layers in IPN by using the FFN. 
To train our model we adopt Adam as the optimizer and use the initial learning rate of $0.0001$ for $30k$ epochs, and gradually reduce it to $0.00001$ for another $20k$ epochs. we set the batch size to $8$ and momentum to $0.9$.

\begin{figure}[t]
 \centering
 \includegraphics[width=0.48\textwidth]{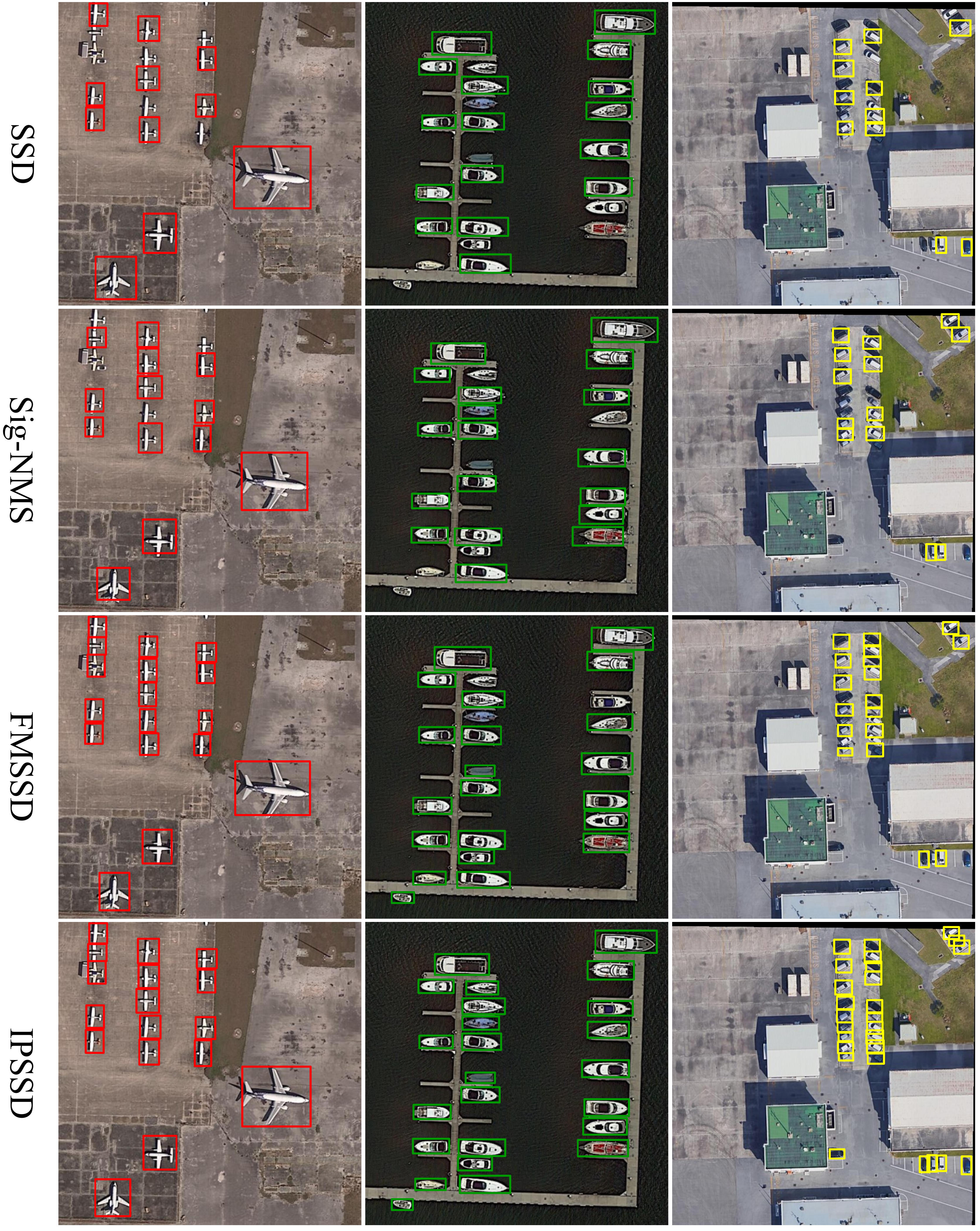}
\caption{Qualitative detection results comparison on small objects of DOTA.}
\label{fig:8}
\end{figure}

\subsection{Model Comparison}
\label{ssec:subhead}

To evaluate the performance of the IPSSD for small object detection, several state-of-the-art models are selected for both quantitative and qualitative comparison.

In Tables \ref{tab:3} and \ref{tab:5}, we report the detection results of our model compare to the other approaches on three small object categories of the DOTA and NWPU datasets. The detection rate of SSD on the DOTA is $70.72$ mAP while processing at $64$ FPS. FMSSD \cite{wang2019fmssd} achieves the detection rates of $78.06$ mAP while processing at $22$ FPS. However, IPSSD achieves $79.24$ mAP detection rate while processing at $53$ FPS. In Fig. \ref{fig:8} we evaluate the performance of IPSSD compare to the other approaches.As the results show, IPSSD can stably produces precise results. Our model also on the NWPU dataset outperforms the state-of-the-art models. Our detector achieves 93.35\% mAP detection rate.
This progress is result of the following components. 
\begin{itemize} 
\item[1)] By combining the IPN into the SSD, we create an architecture where each image scale is featurized and resulted in improving the performance of our detector. 
\item[2)] The FFN can enhance the attention of our proposed model to the whole object parts, which resulted in more accurate small object detection.
\end{itemize}


\begin{table}
\centering
  \caption{Comparison of the performance for small object detection on the test set of DOTA for HBB task.}
  \label{tab:3}
  \begin{tabular}{m{5.3em}|  m{0.7cm} m{0.7cm} m{0.9cm} | m{0.7cm} m{0.4cm}}     \hline
    Methods &  Plane&SV&Ship&mAP&FPS  \\     \hline

{SSD \cite{liu2016ssd}}
 &	                    79.64&62.13&70.41&70.72	& {\textbf{64}} \\ \hline
{Sig-NMS \cite{dong2019sig}}
&	                          86.97&66.19&74.33& 75.83	& 31 \\\hline
{FMSSD \cite{wang2019fmssd}}
       &\textbf {89.14}&68.32&76.73&	78.06&22\\ \hline
{IPSSD} &	{{89.09}}&{\textbf{71.39}}& \textbf {77.26}& 	{\textbf{79.24}}	& {53} \\
        \hline
\end{tabular}
\end{table}
\begin{table}
\centering
  \caption{Comparison of the performance for small object detection on the test set of NWPU.}
  \label{tab:5}
   \begin{tabular}{m{5.3em}| p{0.7cm} p{0.7cm} p{0.9cm}| p{0.7cm} p{0.4cm}}     \hline
    Methods &Plane&Ship&Vehicle& mAP & FPS  \\     \hline
 
{SSD \cite{liu2016ssd}}  &85.16&76.51&62.14& 74.59 & {\textbf{64}}  \\ \hline

{Sig-NMS \cite{dong2019sig}}  &90.94&81.03&78.12& 83.36 & 31 \\   \hline
{FMSSD \cite{wang2019fmssd}} &\textbf {99.72}&89.90&88.23& 92.61 & 22 \\ \hline

{IPSSD} &{{99.63}}&\textbf{91.07}&{\textbf{89.35}}& {\textbf{93.35}} & 53\\ \hline

\end{tabular}
\end{table}

\begin{figure}
 \centering
 \includegraphics[width=0.5\textwidth]{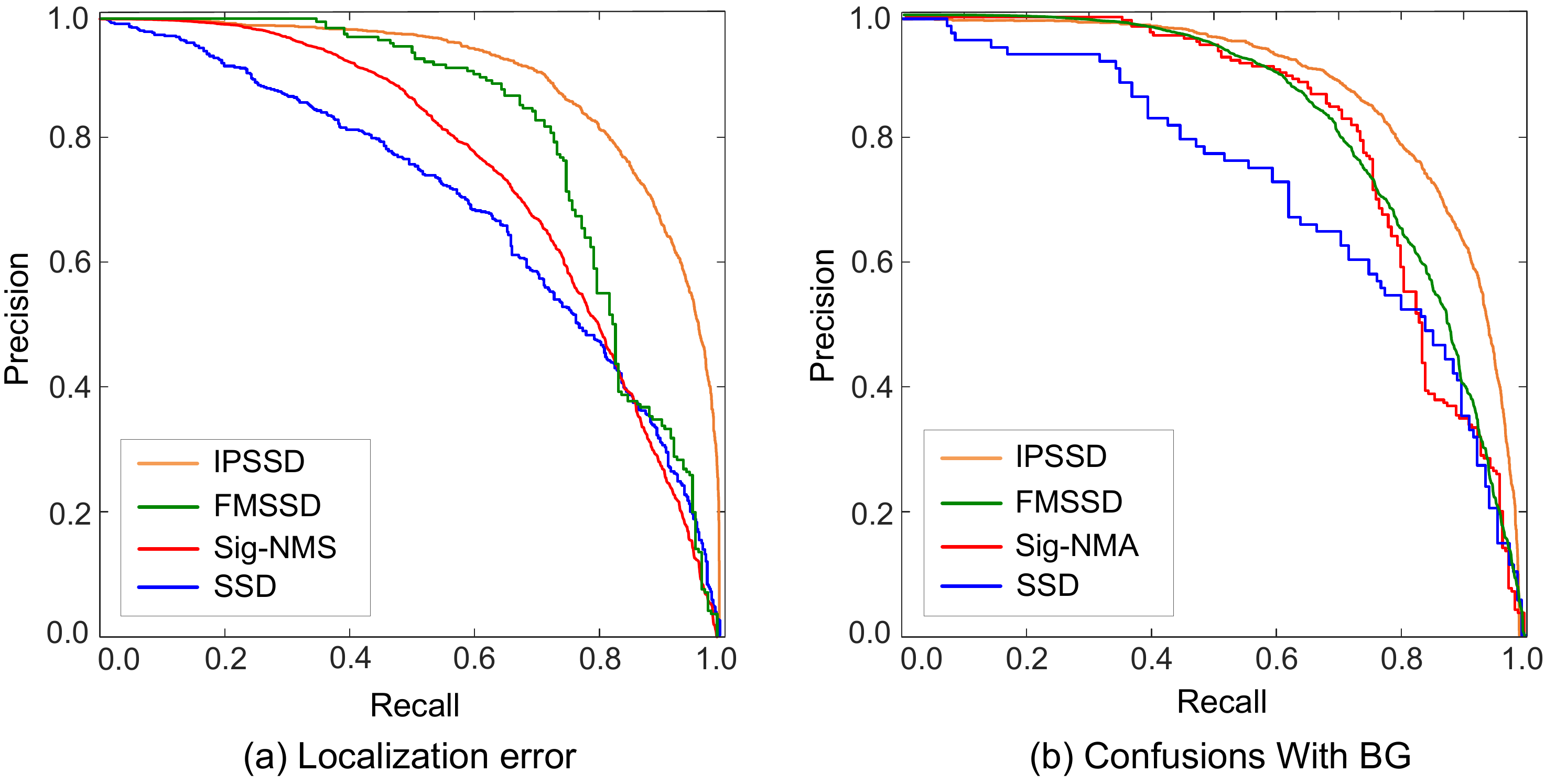}
\caption{{Performance evaluation.}}
\label{fig:ROC}
\end{figure}

Fig. \ref{fig:ROC} shows the curve plot of mean localization error and confusions with background on the DOTA. As demonstrated, IPSSD has better performance in comparison with the other models.

\section{Conclusion}
\label{sec:5}
In this paper, we proposed an effective architecture by adopting image pyramid network into SSD to extract more semantic features for small target detection in RSIs. We conducted extensive experiments on two public datasets and the results show that our model performs better than the other state-of-the-art approaches for detecting small objects. 

\bibliographystyle{IEEEtran}
\bibliography{IEEEexample}

\end{document}